\documentclass[11pt]{article}
\usepackage{amsmath, amssymb, amsthm}
\usepackage{graphicx}
\usepackage{booktabs}
\usepackage{hyperref}
\usepackage{geometry}
\usepackage{float}

\usepackage[numbers,sort&compress]{natbib} 

\title{An Information-Theoretic Framework for Credit Risk Modeling:\\
Unifying Industry Practice with Statistical Theory for Fair and Interpretable Scorecards}
\author{
Agus Sudjianto$^{1,2}$ \\
$^1$H2O.ai, \texttt{agus.sudjianto@h2o.ai} \\
$^2$Center for Trustworthy AI Through Model Risk Management, \\
University of North Carolina Charlotte \\
\\
Denis Burakov$^{3}$ \\
$^3$Amazon, xRiskLab \\
\texttt{contact@xrisklab.ai}
}
\date{\today}

\begin{document}
\maketitle

\begin{abstract}
Credit risk modeling has long relied on Weight of Evidence (WoE) and Information Value (IV) for feature engineering and variable selection, while using the Population Stability Index (PSI) for monitoring model drift. Despite their widespread adoption, the theoretical foundations connecting these industry-standard metrics have remained unclear. This paper establishes a unified information-theoretic framework that reveals these constructs as specific instantiations of classical information divergences. We prove that the industry-standard IV is exactly equivalent to PSI (Jeffreys divergence) computed between good and bad credit outcomes over identical bins. We further introduce standard errors for IV and PSI, bridging information theory with statistical inference by deriving uncertainty bounds through the delta method applied to WoE transformations. Building on this foundation, we formalize the performance--fairness trade-off inherent in credit modeling: maximizing IV with respect to default outcomes (predictive power) while minimizing IV with respect to protected attributes (fairness), leveraging statistical significance testing and probabilistic constraints. Our framework operationalizes this trade-off through automated binning using depth-1 XGBoost stumps and compares three encoding strategies: logistic regression with one-hot encoding, WoE transformation, and constrained XGBoost. The introduction of IV standard errors enables, for the first time, formal hypothesis testing for feature significance and probabilistic fairness constraints, replacing arbitrary thresholds with principled statistical inference. Empirical validation on a realistic credit dataset demonstrates that all approaches achieve comparable predictive performance (AUC 0.82--0.84), confirming that optimal binning derived from information-theoretic principles is more critical than the specific encoding strategy. We implement mixed-integer programming to trace Pareto-efficient solutions along the performance-fairness frontier supported by uncertainty quantification. The unified framework offers both rigorous theoretical justification for widely-used industry practices and practical methodologies for balancing predictive accuracy with fairness considerations in regulated financial environments, while providing the first statistical inference framework for these fundamental credit risk metrics.
\end{abstract}

\section{Introduction}

Credit risk modeling represents one of the most mature applications of statistical learning in finance, with practices codified through decades of regulatory oversight and industry standardization. At the heart of this discipline lie three fundamental constructs: Weight of Evidence (WoE) for feature transformation, Information Value (IV) for variable selection, and Population Stability Index (PSI) for model monitoring \citep{siddiqi2005scorecards, anderson2007toolkit, thomas2017credit, hand1997statistical}. These metrics have become so deeply embedded in credit risk practice that they are often treated as axiomatic, with their theoretical foundations remaining largely implicit.

Parallel to this industry evolution, information theory has provided rigorous mathematical frameworks for measuring distributional differences through divergences such as Kullback--Leibler (KL) and Jensen--Shannon (JS) \citep{kullback1951information, lin1991divergence, endres2003new, cover2005elements}. While these theoretical constructs appear in academic literature, their connections to practical credit modeling have remained underexplored, creating an unfortunate divide between statistical theory and industry application.

This paper bridges that divide by establishing a unified information-theoretic foundation for credit risk modeling. Our key theoretical contribution demonstrates that industry-standard metrics are not ad hoc constructs but rather specific instantiations of well-established information divergences. Most notably, we prove that the IV formula universally used for variable selection is mathematically identical to PSI (Jeffreys divergence) computed between good and bad credit outcomes over the same binning scheme.

Building on this theoretical unification, we make a second novel contribution by introducing standard errors for Information Value and Population Stability Index. This aims to bridge information-theoretic foundations with statistical inference, transforming deterministic industry metrics into proper statistical quantities with uncertainty bounds. By leveraging the relationship between WoE and bin-level log-odds ratios, we derive closed-form expressions for IV standard errors, enabling hypothesis testing, confidence intervals, and probabilistic fairness constraints.

This unification yields immediate practical benefits. By recognizing IV and PSI as information divergences with quantifiable uncertainty, we can leverage both the rich theoretical properties of these measures and rigorous statistical inference to address contemporary challenges in credit modeling, particularly the growing emphasis on fairness and responsible AI. We formalize the performance--fairness trade-off as competing information-theoretic objectives: maximizing IV with respect to default outcomes (predictive power) while minimizing IV with respect to protected demographic attributes (fairness) leveraging uncertainty-aware optimization.

Our empirical contributions validate this framework through comprehensive experiments on realistic credit data. We demonstrate that multiple encoding strategies---logistic regression with one-hot encoding, WoE transformation, and constrained XGBoost---achieve comparable predictive performance when built on the same information-theoretic foundation. This finding suggests that optimal binning derived from information-theoretic principles is more crucial than the specific modeling approach, providing practitioners with flexibility in implementation while maintaining theoretical rigor.

The paper is structured to serve both theoretical and practical needs. Section 2 establishes the information-theoretic foundations, formally defining KL, PSI, and Jensen--Shannon divergences and their relationships. Section 3 demonstrates the equivalence between industry IV and theoretical PSI. Section 4 introduces our novel contribution of IV standard errors, bridging information theory with statistical inference. Section 5 compares different symmetric extensions of KL divergence, while Section 6 formalizes the performance--fairness trade-off using this unified framework enhanced with uncertainty quantification. Sections 7 and 8 provide empirical validation through binning strategy comparisons and bi-objective optimization, respectively. We conclude with implications for both research and practice in credit risk modeling.

\section{Information-Theoretic Foundations}

Information theory provides a principled mathematical framework for quantifying the difference between probability distributions. This section establishes the key divergence measures that will form the foundation of our unified approach to credit risk modeling.

\subsection{Kullback--Leibler Divergence}

The Kullback--Leibler (KL) divergence serves as the fundamental building block for measuring distributional differences. For discrete distributions $P=(p_i)$ and $Q=(q_i)$ defined on the same support, the KL divergence is:

\begin{equation}
D_{\mathrm{KL}}(P\|Q) = \sum_i p_i \log\frac{p_i}{q_i}
\end{equation}

KL divergence quantifies the expected logarithmic difference between distributions $P$ and $Q$, measured according to distribution $P$. A crucial property is its asymmetry: $D_{\mathrm{KL}}(P\|Q) \neq D_{\mathrm{KL}}(Q\|P)$ in general. This asymmetry, while mathematically natural, can complicate practical applications where symmetric measures are preferred.

In the context of credit risk, KL divergence naturally measures how much the distribution of a feature differs between good and bad borrowers. A feature with identical distributions across credit outcomes would have $D_{\mathrm{KL}} = 0$, indicating no predictive value, while features with markedly different distributions would exhibit large KL divergence values.

\subsection{Population Stability Index as Jeffreys Divergence}

The asymmetry limitation of KL divergence motivates the consideration of symmetric alternatives. The Population Stability Index (PSI), ubiquitous in credit risk monitoring, is defined between distributions $P$ and $Q$ as:

\begin{equation}
\label{eq:psi}
\mathrm{PSI}(P,Q) = \sum_i (p_i-q_i)\log\frac{p_i}{q_i}
\end{equation}

Algebraic manipulation reveals that PSI is exactly the Jeffreys divergence:

\begin{equation}
\mathrm{PSI}(P,Q) = D_{\mathrm{KL}}(P\|Q) + D_{\mathrm{KL}}(Q\|P)
\end{equation}

This equivalence establishes PSI as a symmetric extension of KL divergence, addressing the asymmetry concern while maintaining the additive structure that makes it interpretable in credit applications. The Jeffreys divergence possesses several advantageous properties: it is symmetric, always non-negative, and equals zero if and only if $P = Q$.

In credit risk practice, PSI is primarily used for monitoring population drift between development and production samples. The connection to information theory provides theoretical justification for the empirical thresholds commonly employed: PSI values below 0.02 indicating stable populations, values between 0.02 and 0.10 suggesting moderate drift, and values above 0.10 signaling significant population changes requiring model recalibration \citep{pruitt2010psi}.

\subsection{Jensen--Shannon Divergence}

While PSI provides symmetry, it suffers from unboundedness and extreme sensitivity when distribution supports have minimal overlap. The Jensen--Shannon (JS) divergence offers an alternative symmetric extension with more favorable numerical properties.

Given distributions $P$ and $Q$, let $M = \frac{1}{2}(P+Q)$ denote their average. The JS divergence is:

\begin{equation}
\mathrm{JS}(P,Q) = \frac{1}{2}D_{\mathrm{KL}}(P\|M) + \frac{1}{2}D_{\mathrm{KL}}(Q\|M)
\end{equation}

JS divergence possesses several attractive properties absent in PSI. It is bounded above by $\log 2$ (with natural logarithms), making it numerically stable even when distributions have sparse overlap. Additionally, $\sqrt{\mathrm{JS}(P,Q)}$ forms a proper metric, satisfying the triangle inequality and enabling principled comparisons across different variables or time periods.

The bounded nature of JS divergence makes it particularly suitable for applications requiring robust numerical behavior, though it sacrifices some of the extreme sensitivity that makes PSI valuable for drift detection in stable environments.

\section{Weight of Evidence and Information Value: Connecting Industry Practice to Theory}

Credit risk modeling has historically relied on two fundamental constructs for variable transformation and selection. This section demonstrates how these industry-standard practices emerge naturally from information-theoretic principles.

\subsection{Weight of Evidence Transformation}

Weight of Evidence provides a principled method for transforming categorical or binned continuous variables into a form that directly reflects their predictive relationship with credit outcomes. For a variable $X$ partitioned into bins indexed by $j$, let $P_g(j)$ and $P_b(j)$ denote the distributions of good and bad borrowers across bins, respectively. The bin-level Weight of Evidence is:

\begin{equation}
\mathrm{WoE}_j = \log\frac{P_g(j)}{P_b(j)}
\end{equation}

The WoE transformation has several appealing properties for credit modeling. First, it directly represents the log-odds ratio for each bin, providing an intuitive measure of relative credit risk. Second, it transforms categorical variables into continuous scales suitable for linear modeling approaches. Third, bins with identical good/bad distributions receive identical WoE values, naturally capturing the notion of equivalent risk, while a WoE of 0 represents the population baseline risk.

From an information-theoretic perspective, WoE represents the bin-level contribution to the overall KL divergence between good and bad distributions. This connection foreshadows the relationship between WoE and Information Value that we establish in the following subsection.

\subsection{Information Value: The Hidden Jeffreys Divergence}

Information Value aggregates bin-level WoE contributions to produce a single measure of a variable's predictive strength. The industry-standard IV formula, appearing in countless risk management textbooks and regulatory guidance documents, is:

\begin{equation}
\label{eq:iv}
\mathrm{IV} = \sum_j \big(P_g(j)-P_b(j)\big)\log\frac{P_g(j)}{P_b(j)}
\end{equation}

This formula, while widely used, has historically lacked clear theoretical justification. Our key theoretical contribution demonstrates that this industry formula is mathematically identical to the Jeffreys divergence between good and bad distributions over the same binning scheme.

To establish this equivalence, we expand the IV formula:

\begin{align}
\mathrm{IV} &= \sum_j \big(P_g(j)-P_b(j)\big)\log\frac{P_g(j)}{P_b(j)} \\
&= \sum_j P_g(j)\log\frac{P_g(j)}{P_b(j)} - \sum_j P_b(j)\log\frac{P_g(j)}{P_b(j)} \\
&= D_{\mathrm{KL}}(P_g\|P_b) + D_{\mathrm{KL}}(P_b\|P_g) \\
&= \mathrm{PSI}(P_g,P_b)
\end{align}

This proof establishes that \textbf{Information Value is exactly the Population Stability Index (Jeffreys divergence)} computed between good and bad borrower distributions over identical bins. This equivalence provides rigorous theoretical foundation for IV's role in variable selection while connecting it directly to the PSI measures used for model monitoring.

The practical implications are significant. The empirical thresholds for IV interpretation---values below 0.02 indicating weak predictive power, 0.02--0.10 suggesting medium strength, and 0.10--0.30 representing strong discriminatory ability---now inherit the theoretical properties of Jeffreys divergence. Moreover, this connection enables the extension of IV-based approaches to fairness applications, as we demonstrate in subsequent sections.

It should be noted that while these thresholds are based on extensive empirical validation, extremely high IV values can be artificially achieved through overly granular binning schemes that create bins with very few observations. In such cases, IV may primarily capture sampling noise rather than genuine predictive signal. Therefore, careful inspection of bin sample sizes and statistical significance (using the standard errors introduced in the following section) is essential for proper application of this measure in practice.

\section{Statistical Inference for Information Value: Bridging Theory with Uncertainty Quantification}

While the information-theoretic foundation provides rigorous mathematical justification for industry practices, practical applications require uncertainty quantification to enable statistical inference. This section introduces standard errors for Information Value (and hence Population Stability Index), transforming these measures into proper statistical quantities with confidence bounds and hypothesis testing capabilities.

\subsection{The Statistical Nature of WoE and IV}

The key insight enabling statistical inference lies in recognizing that Weight of Evidence represents a centered version of the bin-specific log-odds ratio. Mathematically, WoE is the difference between bin-specific log-odds and the population baseline log-odds:

\begin{equation}
\mathrm{WoE}_j = \underbrace{\log\frac{n_{j,g}}{n_{j,b}}}_{\text{bin log-odds}} - \underbrace{\log\frac{n_g}{n_b}}_{\text{population log-odds}}
\end{equation}

This can be equivalently written in the standard industry form:
\begin{equation}
\mathrm{WoE}_j = \log\frac{n_{j,g}/n_g}{n_{j,b}/n_b} = \log\frac{P_g(j)}{P_b(j)}
\end{equation}

The centering operation is crucial for statistical inference because it preserves the variance structure while providing interpretable baseline comparisons. Since subtracting a constant (the population log-odds) does not affect variance, we have:

\begin{equation}
\mathrm{Var}(\mathrm{WoE}_j) = \mathrm{Var}\left(\log\frac{n_{j,g}}{n_{j,b}}\right)
\end{equation}

This formulation clarifies why a WoE of 0 represents population baseline risk---it occurs when the bin's log-odds equals the overall population log-odds, indicating no deviation from the expected risk level. Since Information Value is a weighted sum of WoE values:

\begin{equation}
\mathrm{IV} = \sum_{j=1}^{J} w_j \cdot \mathrm{WoE}_j
\end{equation}

where $w_j = (p_{b,j} - p_{g,j})$ are the bin-specific weights, we can apply the delta method to propagate uncertainty from individual WoE estimates to the aggregate IV measure.

\subsection{Standard Error Derivation}

The standard error of WoE follows from the delta method applied to the log-odds ratio. The key insight is that WoE inherits the variance of the bin-specific log-odds because the population log-odds is a fixed constant computed from the entire dataset. Applying the delta method to the bin-specific log-odds $\log(n_{j,g}/n_{j,b})$ yields:

\begin{equation}
\mathrm{Var}(\mathrm{WoE}_j) = \mathrm{Var}\left(\log\frac{n_{j,g}}{n_{j,b}}\right) = \frac{1}{n_{j,g}} + \frac{1}{n_{j,b}}
\end{equation}

This can be equivalently expressed in terms of bin size and event rate:
\begin{equation}
\mathrm{Var}(\mathrm{WoE}_j) = \frac{1}{n_j \cdot p_j \cdot (1-p_j)}
\end{equation}

where $n_j = n_{j,g} + n_{j,b}$ is the total bin size and $p_j = n_{j,b}/n_j$ is the bin-specific event rate.

This variance formula has a direct connection to logistic regression: the standard errors of WoE-transformed variables exactly match the standard errors of the corresponding coefficients in a logistic regression model, providing theoretical validation for the widespread use of WoE in credit scoring applications.

For the Information Value, assuming independence across bins (valid for distinct, non-overlapping bins), the variance becomes:

\begin{equation}
\mathrm{Var}(\mathrm{IV}) = \sum_{j=1}^{J} w_j^2 \cdot \mathrm{Var}(\mathrm{WoE}_j)
\end{equation}

Therefore, the standard error of Information Value is:

\begin{equation}
\label{eq:iv_se}
\mathrm{SE}(\mathrm{IV}) = \sqrt{\sum_{j=1}^{J} (p_{b,j} - p_{g,j})^2 \cdot \left(\frac{1}{n_{j,g}} + \frac{1}{n_{j,b}}\right)}
\end{equation}

\subsection{Statistical Significance Testing}

The availability of standard errors enables formal hypothesis testing for Information Value (and PSI). The null hypothesis of no predictive power corresponds to:

\begin{align}
H_0&: \mathrm{IV} = 0 \text{ (no discriminatory power)} \\
H_1&: \mathrm{IV} > 0 \text{ (predictive value exists)}
\end{align}

Under the null hypothesis, the test statistic follows an asymptotic normal distribution:

\begin{equation}
Z = \frac{\mathrm{IV}}{\mathrm{SE}(\mathrm{IV})} \sim \mathcal{N}(0,1)
\end{equation}

This enables p-value calculation and statistical significance testing, replacing arbitrary IV thresholds with principled statistical tests.

\subsection{Confidence Intervals and Practical Implications}

Standard errors also enable construction of confidence intervals for IV estimates:

\begin{equation}
\mathrm{CI}_{1-\alpha}(\mathrm{IV}) = \mathrm{IV} \pm z_{\alpha/2} \cdot \mathrm{SE}(\mathrm{IV})
\end{equation}

These confidence bounds have an immediate appeal in practical applications:

\begin{itemize}
\item \textbf{Feature selection}: Compare confidence intervals rather than point estimates to identify truly significant predictors based on IV
\item \textbf{Model monitoring}: Detect statistically significant changes in IV over time, distinguishing genuine drift from sampling variation
\item \textbf{Fairness assessment}: Quantify uncertainty in demographic IV measures, enabling probabilistic fairness constraints
\end{itemize}

\subsection{Extension to Population Stability Index}

Since we established that $\mathrm{IV} = \mathrm{PSI}$ for identical binning schemes, the same standard error formulation applies directly to PSI calculations. This provides, for the first time, rigorous uncertainty quantification for population stability monitoring, enabling statistical tests for drift detection rather than relying solely on empirical thresholds.

The statistical framework transforms PSI from a deterministic monitoring tool into a proper statistical test, where drift detection becomes a hypothesis testing problem with controlled Type I and Type II error rates.

\section{Symmetric KL Divergences: Comparing PSI and Jensen--Shannon}

Both PSI and Jensen--Shannon divergence represent symmetric extensions of the fundamental KL divergence, but they exhibit markedly different behaviors that affect their suitability for different credit modeling applications.

\subsection{Sensitivity and Boundedness Properties}

The most critical difference lies in their numerical behavior. PSI, being the direct sum of two KL divergences, is unbounded and can exhibit extreme sensitivity when distributions have minimal overlap. In credit applications, this manifests when certain demographic groups or risk segments have very different feature distributions, potentially causing PSI values to explode even for modest population shifts.

Consider a scenario where a particular income bracket contains 10\% of good borrowers but only 0.1\% of bad borrowers in the development sample. If this proportion shifts to 8\% and 0.5\% respectively in production, the PSI contribution from this single bin becomes substantial, potentially dominating the overall measure and triggering false alarms about population stability.

In contrast, Jensen--Shannon divergence is bounded above by $\log 2$, ensuring numerical stability even in extreme cases. This boundedness comes at the cost of reduced sensitivity to population shifts, which may be disadvantageous for drift detection applications where extreme sensitivity is actually desired.

\subsection{Interpretability and Practical Considerations}

From a practitioner's perspective, PSI benefits from decades of institutional knowledge about appropriate threshold values and their business implications. The widely-accepted ranges (0.02, 0.10, 0.30) for PSI interpretation reflect extensive empirical validation across different credit portfolios and economic conditions.

Jensen--Shannon divergence, while theoretically superior in many respects, lacks this institutional context. However, its metric properties ($\sqrt{\mathrm{JS}}$ forms a proper distance metric) enable more principled statistical approaches, such as hypothesis testing for distributional differences or clustering algorithms for market segmentation.

\subsection{Practical Recommendations}

The choice between PSI and JS divergence depends on the specific application requirements:

\begin{itemize}
\item \textbf{Model monitoring}: PSI's sensitivity makes it preferable for detecting population drift, where false negatives (missing actual drift) are more costly than false positives.
\item \textbf{Variable selection}: Both measures perform comparably, though JS divergence's boundedness may be preferable in automated model development pipelines.
\item \textbf{Fairness applications}: JS divergence's stability and metric properties make it more suitable for optimization algorithms and comparative analyses across protected groups.
\end{itemize}

\section{Formalizing the Performance--Fairness Trade-off with Statistical Inference}

The unified information-theoretic framework, enhanced with uncertainty quantification, enables a precise formalization of the performance--fairness trade-off that has become central to responsible AI in financial services. By recognizing that both predictive power and demographic fairness can be measured using the same mathematical constructs with quantifiable uncertainty, we can establish explicit optimization objectives that balance these competing goals while accounting for statistical significance and confidence bounds.

\subsection{Dual Information Value Objectives}

Consider a credit modeling scenario with target variable $Y \in \{0,1\}$ representing non-default (0) and default (1) outcomes, and protected attribute $A \in \{0,1\}$ indicating membership in a protected demographic group. For any candidate feature $X$, we can compute two distinct IV measures using identical mathematical formulations but different conditioning variables:

\begin{align}
\text{Predictive IV:}\quad \mathrm{IV}_{\text{perf}}(X) &= D_{\mathrm{KL}}(P_{X|Y=0}\|P_{X|Y=1}) + D_{\mathrm{KL}}(P_{X|Y=1}\|P_{X|Y=0}) \\
\text{Fairness IV:}\quad \mathrm{IV}_{\text{fair}}(X) &= D_{\mathrm{KL}}(P_{X|A=0}\|P_{X|A=1}) + D_{\mathrm{KL}}(P_{X|A=1}\|P_{X|A=0})
\end{align}

The first measure quantifies how much feature $X$ differentiates between good and bad credit outcomes---precisely the discriminatory power that credit models seek to maximize. The second measure quantifies how much the same feature differentiates between protected and reference demographic groups---exactly the discriminatory behavior that fairness regulations seek to minimize.

\subsection{The Fundamental Trade-off}

This dual formulation crystallizes the performance--fairness trade-off:

\begin{itemize}
\item \textbf{Performance objective}: Maximize $\mathrm{IV}_{\text{perf}}(X)$ to achieve strong separation between good and bad borrowers
\item \textbf{Fairness objective}: Minimize $\mathrm{IV}_{\text{fair}}(X)$ to ensure weak separation between protected and reference groups
\end{itemize}

The mathematical equivalence of these objectives---both computed as Jeffreys divergences over different partitions---creates a fundamental tension. Features that strongly differentiate credit outcomes often correlate with demographic characteristics, making simultaneous optimization challenging.

\subsection{Binning as a Fairness Regularizer}

The information-theoretic framework reveals how modeling choices affect the performance--fairness balance. Binning granularity serves as a critical control mechanism: finer binning typically increases both $\mathrm{IV}_{\text{perf}}$ and $\mathrm{IV}_{\text{fair}}$, enhancing predictive power while potentially exacerbating fairness concerns.

This insight suggests that limiting binning depth or applying regularization acts as an implicit fairness regularizer by constraining information leakage about protected attributes. The framework thus provides principled guidance for practitioners seeking to balance these competing objectives:

\begin{enumerate}
\item \textbf{Coarse binning}: Reduces both predictive power and potential bias, suitable for highly regulated environments
\item \textbf{Adaptive binning}: Adjusts granularity based on demographic composition within bins, optimizing the trade-off
\item \textbf{Constrained binning}: Explicitly incorporates fairness constraints into the binning optimization process
\end{enumerate}

\subsection{Uncertainty-Aware Fairness Constraints}

The introduction of IV standard errors transforms deterministic fairness constraints into probabilistic assessments. Rather than hard thresholds of the form $\mathrm{IV}_{\text{fair}}(X) \leq \epsilon$, we can now implement uncertainty-aware constraints:

\begin{equation}
P(\mathrm{IV}_{\text{fair}}(X) \leq \epsilon) \geq \alpha
\end{equation}

where $\alpha$ represents the confidence level (e.g., 95\%). This probabilistic formulation accounts for estimation uncertainty and provides more robust fairness assessments.

For practical implementation, this constraint can be approximated using the normal distribution:

\begin{equation}
\mathrm{IV}_{\text{fair}}(X) + z_{\alpha} \cdot \mathrm{SE}(\mathrm{IV}_{\text{fair}}(X)) \leq \epsilon
\end{equation}

This approach transforms binary pass/fail fairness decisions into risk-based assessments with quantified confidence levels, enabling more nuanced regulatory compliance strategies.

\vspace{0.5em}
\noindent\textbf{Practical Example: From Binary to Probabilistic Assessment.}\par
\vspace{0.3em}

Consider a gender feature with $\mathrm{IV}_{\text{fair}} = 0.066$ and $\mathrm{SE}(\mathrm{IV}_{\text{fair}}) = 0.0069$, evaluated against a fairness threshold of $\epsilon = 0.05$:

\begin{itemize}
\item \textbf{Traditional approach}: ``Gender IV $= 0.066 \mathbin{>} 0.05$ threshold'' $\rightarrow$ Violation
\item \textbf{Uncertainty-aware approach}: ``$P(\text{Gender IV} > 0.05)=98.9\%$'' $\rightarrow$ High confidence of violation
\end{itemize}

The probabilistic assessment provides richer guidance for decision-making. A violation with $98.9\%$ confidence constitutes strong evidence of bias and would typically justify remediation or feature mitigation, whereas a violation with only $55\%$ confidence may motivate additional data collection, sensitivity analysis, or feature engineering before deciding whether feature exclusion or adjustment is appropriate.

\subsection{Connection to Established Fairness Metrics}

The information-theoretic formulation connects naturally to established fairness criteria in machine learning. Demographic parity requires $P(\hat{Y}=1|A=0) = P(\hat{Y}=1|A=1)$, which corresponds to minimizing $\mathrm{IV}_{\text{fair}}$ computed on model predictions rather than input features. Equalized odds extends this to condition on true outcomes, relating to IV measures computed within outcome subgroups.

This connection enables practitioners to translate between information-theoretic objectives and regulatory requirements, providing a bridge between statistical optimization and compliance considerations, now enhanced with statistical confidence measures.

\section{Empirical Validation: Binning Strategies and Encoding Approaches}

Having established the theoretical foundation, we now demonstrate how these insights translate into practical modeling decisions. This section evaluates different approaches to operationalizing the binned features that underpin all information-theoretic measures, using a realistic credit risk simulation to validate our framework.

\subsection{Unified Binning Foundation}

All modeling approaches in our framework share a common foundation: optimal binning derived from depth-1 XGBoost decision stumps. This data-driven binning strategy automatically discovers splits that maximize information gain, directly implementing our theoretical objective of maximizing IV for predictive power.

The binning process operates by fitting single-node decision trees (stumps) to each feature, using the Gini impurity criterion to identify splits that best separate good and bad borrowers. This approach offers several advantages:

\begin{enumerate}
\item \textbf{Theoretical consistency}: The information gain objective directly corresponds to maximizing KL divergence between outcome distributions
\item \textbf{Automated optimization}: Removes subjective binning decisions while incorporating domain-appropriate constraints
\item \textbf{Unified foundation}: The same bins serve multiple purposes---IV calculation, WoE transformation, and PSI monitoring
\end{enumerate}

Importantly, the binning granularity (controlled by minimum bin size requirements and maximum number of splits) serves as our primary mechanism for managing the performance--fairness trade-off identified in Section 5.

\subsection{Alternative Encoding Strategies}

Given optimal bins from XGBoost stumps, we compare three distinct strategies for incorporating binned features into predictive models:

\subsubsection{Logistic Regression with One-Hot Encoding (LR-OneHot)}
Each bin becomes a binary indicator variable, preserving the discrete nature of our information-theoretic computations while maintaining complete interpretability. This approach directly models the bin-specific risk contributions without imposing monotonicity assumptions.

\subsubsection{Logistic Regression with Weight of Evidence (LR-WoE)}
Bins are transformed using WoE scores from Equation (4), creating a single continuous feature per variable. This transformation directly embeds the log-odds information used in IV calculations, providing the most direct implementation of our theoretical framework.

\subsubsection{XGBoost with Depth-1 Constraints (XGB-D1)}
The same stumps used for binning are refined through gradient boosting with monotonicity constraints. This approach maintains the binned structure while allowing ensemble refinement, combining interpretability with predictive flexibility.

Each approach leverages identical underlying binning while operationalizing the information differently, enabling isolation of encoding strategy effects from binning quality.

\subsection{Experimental Setup and Dataset}

We evaluate our framework using a synthetic credit risk dataset designed to reflect realistic industry characteristics while enabling controlled experimentation. The dataset contains 20,000 observations with 7 predictive features, 1 protected demographic attribute, and a binary default indicator.\footnote{Complete data and reproducible code are available at: \url{https://github.com/asudjianto-xml/Information-Theoretic-for-Credit-Modeling}}

The feature set includes typical credit risk variables: mortgage amounts, account balances, past due amounts, utilization ratios, delinquency history indicators, recent inquiry flags, and trade count variables. The target variable exhibits a realistic default rate of approximately 15\%, while the protected attribute shows a 30\%/70\% demographic split, creating meaningful conditions for evaluating performance--fairness trade-offs.

\subsection{Results and Analysis}

Table \ref{tab:performance_comparison} summarizes the predictive performance across all three encoding strategies, evaluated using both discriminatory power (AUC) and probability calibration (Log Loss) on independent test data.

\begin{table}[H]
\centering
\small
\begin{tabular}{l|ccc|ccc|ccc}
\toprule
 & \multicolumn{3}{c|}{LR One-Hot} & \multicolumn{3}{c|}{LR (WoE)} & \multicolumn{3}{c}{XGBoost (Depth=1)} \\
Rounds & Train AUC & Test AUC & Gap & Train AUC & Test AUC & Gap & Train AUC & Test AUC & Gap \\
\midrule
10  & 0.818 & 0.826 & -0.007 & 0.813 & 0.819 & -0.006 & 0.816 & 0.821 & -0.005 \\
50  & 0.829 & 0.839 & -0.010 & 0.816 & 0.823 & -0.007 & 0.831 & 0.840 & -0.009 \\
100 & 0.829 & 0.839 & -0.010 & 0.816 & 0.823 & -0.007 & 0.831 & 0.840 & -0.009 \\
\bottomrule
\end{tabular}
\caption{Performance comparison across encoding strategies. Gap = Test AUC - Train AUC. Negative gaps indicate slight generalization improvement, reflecting the regularization benefits of constrained binning.}
\label{tab:performance_comparison}
\end{table}

Several key findings emerge from these results:

\subsubsection{Performance Equivalence Across Encodings}
All three approaches achieve comparable AUC performance (0.82--0.84), confirming our theoretical prediction that optimal binning derived from information-theoretic principles is more critical than the specific encoding strategy. This finding provides practitioners with flexibility in implementation while maintaining performance guarantees.

\subsubsection{WoE Transformation Efficiency}
The LR-WoE approach achieves competitive performance with the most parsimonious representation, using only one transformed feature per original variable. This validates our theoretical insight that WoE captures the essential predictive information contained in each bin's log-odds ratio.

\subsubsection{Regularization Effects}
The consistent negative gaps (0.005--0.012) across all models reflect beneficial regularization from our constrained modeling approaches. Limited binning depth and monotonicity constraints act as implicit regularizers, improving generalization while serving as the fairness control mechanism identified in Section 5.

\subsection{Implications for Practice}

These empirical results validate several important practical principles:

\begin{enumerate}
\item \textbf{Binning quality dominates encoding choice}: Practitioners should focus primarily on developing principled binning strategies rather than optimizing encoding methods.
\item \textbf{WoE provides theoretical and practical advantages}: The direct connection to information theory, combined with computational efficiency, makes WoE transformation particularly attractive.
\item \textbf{Regularization enhances both performance and fairness}: Constraint-based approaches improve generalization while providing fairness control mechanisms.
\item \textbf{Uncertainty-aware trade-off assessment}: The availability of IV standard errors enables conservative performance-fairness evaluation using confidence-adjusted ratios:
\begin{equation}
R_{\text{conservative}} = \frac{\mathrm{IV}_{\text{perf}} - k \cdot \mathrm{SE}(\mathrm{IV}_{\text{perf}})}{\mathrm{IV}_{\text{fair}} + k \cdot \mathrm{SE}(\mathrm{IV}_{\text{fair}})}
\end{equation}
where $k$ represents the desired confidence level (e.g., $k = 1.96$ for 95\% confidence or $k = 2.58$ for 99\% confidence). This approach provides robust trade-off assessment that accounts for estimation uncertainty in both objectives.
\end{enumerate}

It is important to note that this study prioritizes framework validation over absolute performance optimization. We intentionally avoided extensive hyperparameter tuning to focus on demonstrating the consistency and flexibility of our information-theoretic approach across different implementation strategies.

\section{Bi-objective Optimization: Operationalizing the Performance-Fairness Trade-off}

The theoretical framework established in Section 5 requires practical implementation methods for balancing competing performance and fairness objectives. This section presents a mixed-integer programming approach that enables systematic exploration of the Pareto frontier between these dual information value objectives(\citep{pangia2024lda}).

\subsection{Mathematical Formulation}

We implement a constrained optimization framework that integrates fairness considerations directly into model estimation while maintaining interpretability through monotonicity constraints. Let $\{(x_i, y_i, a_i)\}_{i=1}^n$ denote features, default outcomes, and protected group memberships for $n$ observations.

The model takes the form of a logistic scorecard:
\begin{equation}
\hat{p}_i = \sigma\left(\beta_0 + \sum_{j=1}^p \beta_j x_{ij}\right)
\end{equation}
where $\sigma$ denotes the logistic function and $\beta$ represents the coefficient vector to be optimized.

\subsubsection{Predictive Objective}
Model-level predictive power is measured through Information Value computed on model scores rather than individual features:
\begin{equation}
\mathrm{IV}_{\text{model}} = \sum_k \left(P_g(k) - P_b(k)\right) \log\frac{P_g(k)}{P_b(k)}
\end{equation}
where $k$ indexes score bins and $P_g(k), P_b(k)$ represent the distributions of good and bad borrowers across these bins.

\subsubsection{Fairness Constraint}
Demographic fairness is enforced through a constraint on the Information Value computed between protected and reference groups:
\begin{equation}
\mathrm{IV}_{\text{demographic}} = \sum_k \left(P_{a=0}(k) - P_{a=1}(k)\right) \log\frac{P_{a=0}(k)}{P_{a=1}(k)} \leq \epsilon
\end{equation}
where $\epsilon$ represents a fairness budget parameter that controls the maximum allowable demographic differentiation.

\subsubsection{Monotonicity Constraints}
Credit risk models must satisfy regulatory and business requirements for monotonic relationships between risk factors and predicted outcomes. We enforce these through explicit coefficient constraints:
\begin{align}
\beta_j &\geq 0 \quad \forall j \in \mathcal{M}_{+} \text{ (risk-increasing variables)} \\
\beta_j &\leq 0 \quad \forall j \in \mathcal{M}_{-} \text{ (risk-decreasing variables)}
\end{align}
where $\mathcal{M}_{+}$ and $\mathcal{M}_{-}$ denote the sets of variables that should increase and decrease estimated risk, respectively.

\subsubsection{Complete Mixed-Integer Program}
The integrated optimization problem becomes:
\begin{align}
\max_{\beta} \quad & \mathrm{IV}_{\text{model}} \\
\text{subject to} \quad & \mathrm{IV}_{\text{demographic}} \leq \epsilon \\
& \beta_j \geq 0 \quad \forall j \in \mathcal{M}_{+} \\
& \beta_j \leq 0 \quad \forall j \in \mathcal{M}_{-}
\end{align}

This formulation enables systematic exploration of the performance--fairness trade-off by varying the fairness budget $\epsilon$ and solving the resulting constrained optimization problem for each value.

\subsection{Experimental Implementation}

We apply this framework to the credit dataset described in Section 6, preprocessing features using the unified binning approach established previously. The optimization procedure involves two sequential steps:

\begin{enumerate}
\item \textbf{Feature discretization}: Apply depth-1 XGBoost stumps to identify optimal bin boundaries for each variable
\item \textbf{Constrained model fitting}: Solve the mixed-integer program for various values of the fairness budget $\epsilon$
\end{enumerate}

We systematically vary the fairness budget across the range $\epsilon \in \{0.1, 0.3, 0.5, 0.7, 1.0, 1.5, 2.0, 2.5, 3.0\}$, generating solutions that span the complete performance--fairness spectrum. For each configuration, we record both performance metrics (model IV and AUC) and fairness metrics (demographic IV and Adverse Impact Ratio).

\subsection{Results: Tracing the Pareto Frontier}

\begin{figure}[H]
    \centering
    \includegraphics[width=0.75\textwidth]{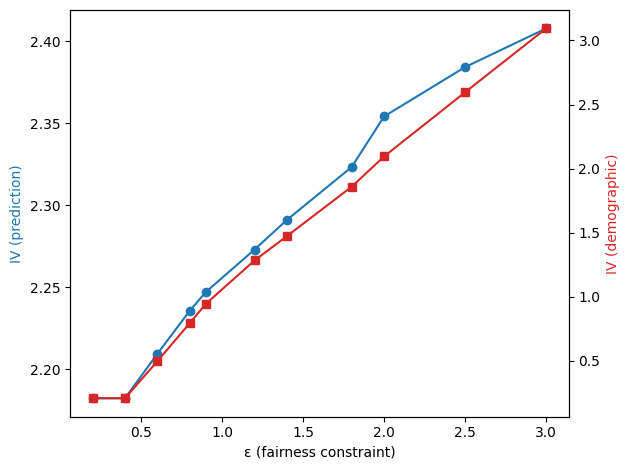}
    \caption{Performance--Fairness Trade-off: Predictive IV (blue, left axis) versus Demographic IV (red, right axis) across fairness budget values. The inverse relationship demonstrates the fundamental tension between predictive power and demographic fairness.}
    \label{fig:pareto_iv}
\end{figure}

\begin{figure}[H]
    \centering
    \includegraphics[width=0.75\textwidth]{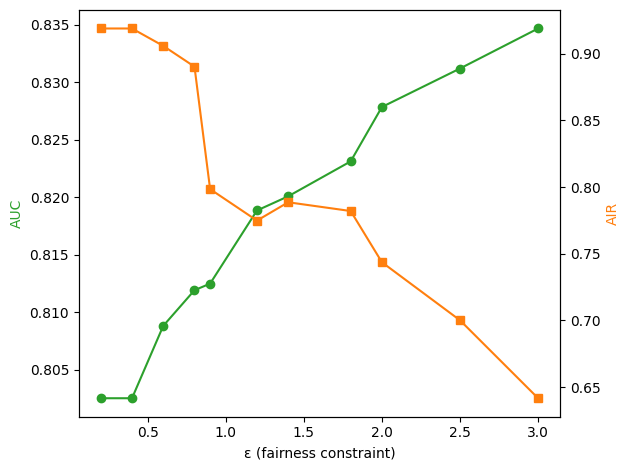}
    \caption{Practical Fairness Metrics: AUC (green, left axis) versus Adverse Impact Ratio (orange, right axis) across fairness budget values. Higher AIR values indicate greater fairness, while AUC measures discriminatory performance.}
    \label{fig:pareto_performance}
\end{figure}

Figures \ref{fig:pareto_iv} and \ref{fig:pareto_performance} display the resulting Pareto frontiers, demonstrating the systematic trade-off between performance and fairness objectives.

\begin{figure}[H]
    \centering
    \includegraphics[width=0.6\textwidth]{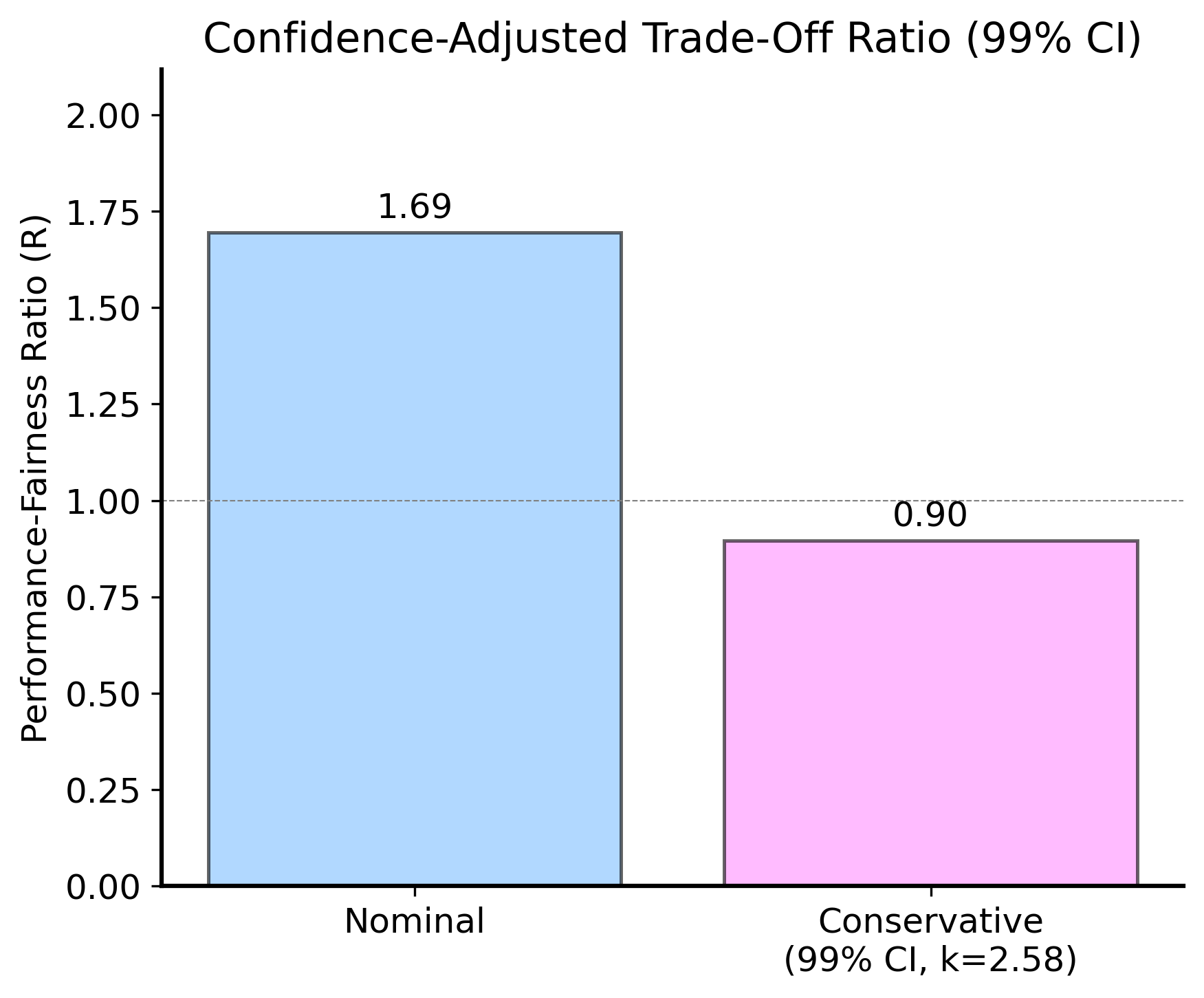}
    \caption{Confidence-Adjusted Trade-Off Ratio (99\% CI) for the \texttt{Mortgage} feature. The conservative ratio accounts for estimation uncertainty in both performance and fairness IVs. At the 99\% confidence level, the ratio drops below 1.0, indicating that we can no longer assert with high confidence that performance gains outweigh fairness costs.}
    \label{fig:conservative_ratio}
\end{figure}

Figure~\ref{fig:conservative_ratio} illustrates the impact of uncertainty-aware adjustment on the performance--fairness ratio, demonstrating how conservative estimation (e.g., at the 99\% confidence level) leads to a more risk-averse interpretation of the trade-off.

\subsubsection{Key Findings}

The Pareto frontier analysis reveals several critical insights:

\begin{enumerate}
\item \textbf{Smooth trade-off curve}: The relationship between performance and fairness metrics follows a smooth, predictable curve, enabling systematic decision-making about appropriate operating points.

\item \textbf{Diminishing returns}: Initial fairness improvements can be achieved with minimal performance degradation, but achieving very high fairness standards requires substantial sacrifice in predictive power.

\item \textbf{Practical operating ranges}: Most practical applications will operate in the middle region of the frontier ($\epsilon \in [0.5, 2.0]$), where both performance and fairness objectives receive meaningful consideration.

\item \textbf{Regulatory compliance}: The framework enables systematic compliance with fairness regulations by selecting operating points that meet specific Adverse Impact Ratio thresholds (e.g., AIR $\ge$ 0.8) while maximizing predictive performance within those constraints.
\end{enumerate}

\subsection{Practical Decision Framework}

The Pareto frontier provides decision-makers with a systematic approach to model selection that balances business objectives with regulatory requirements. The process involves several key steps:

\begin{enumerate}
\item \textbf{Regulatory assessment}: Determine minimum fairness requirements based on applicable regulations and institutional risk tolerance.

\item \textbf{Business prioritization}: Establish minimum acceptable performance levels based on business requirements and competitive positioning.

\item \textbf{Operating point selection}: Choose the Pareto-efficient solution that best satisfies both regulatory and business constraints.

\item \textbf{Sensitivity analysis}: Evaluate robustness of the selected operating point to changes in data distribution or regulatory requirements.
\end{enumerate}

\subsection{Computational Considerations}

The mixed-integer programming formulation scales efficiently with problem size, making it suitable for production deployment. Key computational advantages include:

\begin{itemize}
\item \textbf{Convex objective}: The IV maximization objective ensures reliable convergence to global optima
\item \textbf{Linear constraints}: Monotonicity and fairness constraints can be expressed as linear inequalities, enabling efficient solution methods
\item \textbf{Parallelizable}: Different values of $\epsilon$ can be solved independently, enabling parallel computation of the complete Pareto frontier
\end{itemize}

\section{Implications for Credit Risk Practice}

The unified information-theoretic framework presented in this paper has significant implications for both current practices and future developments in credit risk modeling. Our theoretical analysis provides rigorous mathematical justification for industry practices that have evolved through decades of empirical application, while offering new capabilities for addressing contemporary challenges in responsible AI.

The equivalence between Information Value and Jeffreys divergence validates the widespread use of IV for variable selection, providing institutions with principled explanations for modeling choices to regulators and auditors. The connection to established statistical theory increases confidence in existing model development processes while enabling new practitioners to learn industry methods within a coherent theoretical framework rather than as isolated techniques.

The framework enables several practical enhancements to traditional credit risk modeling workflows. The connection between optimal binning and information gain provides algorithmic guidance for feature discretization, replacing subjective binning decisions with principled optimization procedures that improve both model consistency and development efficiency. More importantly, the bi-objective framework enables systematic incorporation of fairness considerations from the earliest stages of model development, rather than as post-hoc adjustments. This integration typically produces better solutions than sequential approaches that optimize performance first and address fairness concerns later.

Recognition that IV and PSI represent the same mathematical construct computed on different data partitions enables integrated monitoring systems that track both predictive degradation and population drift using consistent mathematical principles. This unified approach provides quantitative measures for assessing model risk arising from population shifts, feature instability, and fairness degradation, which can be integrated into existing model risk management frameworks to provide early warning systems for model deterioration.

The information-theoretic perspective also offers new approaches to traditional risk management challenges. Information divergence measures can be used to design stress testing scenarios that systematically explore the impact of distributional shifts on model performance and fairness characteristics. Additionally, the connection between industry metrics and established statistical theory facilitates communication with regulators and enables more sophisticated analyses for regulatory submissions.

For financial institutions, the framework provides both theoretical clarity about existing methods and practical tools for next-generation credit risk models that explicitly account for fairness considerations. The approach maintains the interpretability and regulatory compliance essential in financial services while providing mathematical rigor for algorithmic decision-making in an increasingly regulated environment.

\section{Limitations and Future Research}

While the unified framework provides significant theoretical and practical advances, several limitations suggest important directions for future research. The framework's reliance on binned features, while consistent with industry practice, may sacrifice information contained in continuous distributions. Future work could explore information-theoretic approaches that operate directly on continuous distributions while maintaining interpretability. Additionally, our empirical validation focuses primarily on logistic regression and constrained tree models, and extension to more complex model architectures such as neural networks and ensemble methods while maintaining interpretability represents an important research direction.

The fairness formulation addresses only single protected attributes, while real-world applications often require consideration of intersectional fairness across multiple demographic dimensions. This limitation suggests the need for extensions to multi-dimensional information divergence measures that can handle complex demographic interactions. Current approaches also treat fairness constraints as static requirements, but future research could explore adaptive fairness constraints that adjust based on economic conditions, regulatory changes, or portfolio composition shifts.

The framework suggests several promising research opportunities that could significantly extend its impact. Natural extensions to feature selection algorithms that explicitly balance predictive information with fairness considerations could lead to more robust variable selection procedures. Combining information-theoretic measures with causal inference frameworks could enable discrimination between legitimate predictive relationships and problematic dependencies that arise from societal biases, addressing fundamental questions about the sources of demographic disparities in credit outcomes.

Extension to scenarios with multiple competing objectives—performance, fairness, interpretability, and stability—could provide more comprehensive decision support for complex modeling environments. This multi-objective perspective becomes increasingly important as financial institutions must simultaneously satisfy diverse stakeholder requirements while maintaining competitive performance.

The implications extend beyond credit risk to any domain where practitioners must balance predictive accuracy with fairness considerations. By establishing clear mathematical relationships between these competing objectives, the framework provides a template for responsible model development that can be adapted to diverse applications while maintaining theoretical rigor and practical interpretability. As the financial industry continues to grapple with responsible AI deployment, this information-theoretic foundation provides a sound path forward that respects both predictive performance requirements and fairness imperatives.

\section{Conclusion}

This paper establishes a unified information-theoretic foundation for credit risk modeling that bridges the historical divide between industry practice and statistical theory. Our central theoretical contribution demonstrates that the ubiquitous metrics in credit scoring---Weight of Evidence, Information Value, and Population Stability Index---are principled instantiations of classical information divergences rather than ad hoc constructs.

The mathematical equivalence between Information Value and Jeffreys divergence provides rigorous justification for IV's widespread use in variable selection while connecting it directly to the theoretical properties of symmetric KL divergence. This unification extends naturally to fairness considerations, revealing that the same mathematical framework measuring predictive power can quantify demographic bias, creating an explicit performance--fairness trade-off that practitioners must navigate.

Our empirical validation demonstrates the practical value of this theoretical foundation. The comparison of encoding strategies---one-hot, WoE transformation, and constrained XGBoost---confirms that optimal binning derived from information-theoretic principles is more crucial than specific modeling approaches, with all methods achieving comparable performance (AUC 0.82--0.84). The WoE transformation's efficiency in achieving competitive results with minimal feature complexity directly validates our theoretical insight that WoE captures essential predictive information in each bin.

The mixed-integer programming implementation shows how the framework can be operationalized for responsible AI applications. By systematically tracing the Pareto frontier between performance and fairness metrics, decision-makers can select operating points that align with regulatory requirements and business priorities while maintaining model interpretability through monotonicity constraints.

Beyond immediate practical applications, this work establishes several important research directions. The connection between binning granularity and fairness regularization suggests opportunities for developing adaptive binning strategies that automatically balance competing objectives. The framework's information-theoretic foundation also provides a principled basis for extending to other domains where similar trade-offs between accuracy and fairness arise.

For practitioners, this framework offers both theoretical clarity about existing methods and practical tools for next-generation credit risk models that explicitly account for fairness considerations. The approach maintains the interpretability and regulatory compliance essential in financial services while providing mathematical rigor for algorithmic decision-making in an increasingly regulated environment.

As the financial industry continues to grapple with responsible AI deployment, this information-theoretic foundation provides a sound path forward that respects both predictive performance requirements and fairness imperatives. The unified framework demonstrates that rigorous statistical theory and practical industry needs are not competing objectives but rather complementary aspects of effective credit risk modeling.

The implications extend beyond credit risk to any domain where practitioners must balance predictive accuracy with fairness considerations. By establishing clear mathematical relationships between these competing objectives, the framework provides a template for responsible model development that can be adapted to diverse applications while maintaining theoretical rigor and practical interpretability.

\nocite{*}


\end{document}